\documentclass[preprints,article,accept,moreauthors,pdftex]{Definitions/mdpi} 

\usepackage{lscape}
\graphicspath{{figs/}}
\setitemize{parsep=6pt,itemsep=0pt,leftmargin=*,labelsep=5.5mm}
\setenumerate{parsep=6pt,itemsep=0pt,leftmargin=*,labelsep=5.5mm}
\setlist[description]{itemsep=0mm}
\firstpage{1} 
\pubvolume{xx}
\issuenum{1}
\articlenumber{5}
\pubyear{2020}
\copyrightyear{2020}

\pdfoutput=1

\PassOptionsToPackage{hyphens}{url}
\usepackage{hyperref}
\usepackage{array}
\usepackage{float}
\usepackage{longtable, booktabs}
\usepackage{caption}
\usepackage[para]{threeparttable}
\usepackage[labelformat=simple]{subcaption}

\DeclareCaptionLabelFormat{subcaptionlabel}{\normalfont(\textbf{#2}\normalfont)}
\usepackage{amssymb}
\usepackage{geometry}
\Title{SatImNet: Structured and Harmonised Training Data for Enhanced Satellite Imagery Classification}

\Author{Vasileios Syrris 
 $^{1,}$*\orcidA{}, Ondrej Pesek $^{2}$ and Pierre Soille $^{1}$\orcidB{}}

\AuthorNames{Vasileios Syrris, Ondrej Pesek and Pierre Soille}

\address{%
$^{1}$ \quad {European Commission, Joint Research Centre (JRC), 21027 Ispra,  
 Italy; pierre.soille@ec.europa.eu}\\  
$^{2}$ \quad {Faculty of Civil Engineering, Czech Technical University in Prague, 16636 Prague, Czechia; 
 ondrej.pesek@ext.ec.europa.eu}} 

\corres{Correspondence: vasileios.syrris@ec.europa.eu; Tel.: +39-0332-789525}

\abstract{Automatic supervised classification with complex modelling such as deep neural networks requires the availability of representative training data sets. While there exists a plethora of data sets that can be used for this purpose, they are usually very heterogeneous and not interoperable. In this context, the present work has a twofold objective: i) to describe procedures of open-source training data management, integration, and data retrieval, and ii) to demonstrate the practical use of varying source training data for remote sensing image classification. For the former, we propose {SatImNet}, a collection of open training data, structured and harmonized according to specific rules. For the latter, two modelling approaches based on convolutional neural networks have been designed and configured to deal with satellite image classification and segmentation.}


\begin{document}

\section{Introduction}
\label{sec:intro}
Data-driven modelling requires sufficient and representative samples that capture and convey significant information about the phenomenon under study. Especially in the case of deep and convolutional neural networks (DNN--CNN), the usage of big training sets is a requisite to estimate adequately the high number of model weights (i.e., the strength of the connection between neural nodes) and avoid over-fitting. The lack of sizeable and labelled training data may be addressed by data augmentation \cite{Shorten2019}, deep modelling techniques such as the generative adversarial networks \cite{Howe2019}, transfer learning \cite{Pan2010} and domain adaptation \cite{Wang2018}. Regarding transfer learning, for instance, there exist large collections of pre-trained models dealing with image classification \cite{Hoeser2020, Bianco2018}. However, these models have been trained on true colour images showing humans, animals, or landscapes, and it remains an open question whether transfer learning, without meticulous domain adaptation and fine-tunings, improves the generic purpose classification or segmentation of satellite images with various spectral, spatial, and temporal resolutions.

The collection of good quality training sets for supervised learning is an expensive, error-prone \cite{ball2017}, and time-consuming procedure. It involves manual or semi-automatic label annotation, verification, and deployment of a suitable sampling strategy like systematic, stratified, reservoir, cluster, snowball, time-location, and many other sampling techniques \cite{Thompson2012}. In addition, consideration of sampling and non-sampling errors and biases need to be taken into account and corrected for. In satellite image classification, factors such as the spectral, spatial, radiometric, and temporal resolution, type of sensor (active or passive \cite{Schott1996}), and data processing level (radiometric and geometric calibration, geo-referencing, atmospheric correction), to name a few, synthesise a manifold of concepts and features that need to be accounted for.

On the other hand, deep supervised learning as well as complex and multi-parametric modelling require big training sets.  A potential solution to this issue is the collection of different training data sets and their exploitation in such a way that they complement each other and act in a synergistic manner.  However, the joint use of two or more existing training data sets require careful examination of their individual features and organisation. To ease this process, we propose a methodology to organise open and freely available training data sets designed for satellite image classification in view of fusing them with other Earth observation (EO)-based products. The proposed methodology is then applied to a series of seven free and open data sets and leads to the {SatImNet} collection. The process of information retrieval has been optimized over a distributed disk storage system. We also demonstrate the blending of different information layers by exploiting deep neural network modelling approaches with the objective to solve concurrently the tasks of image classification and semantic segmentation. This work ties in with the framework of the European strategy for open science \cite{eu2020} which strives to make research more open, global, collaborative, reproducible, and verifiable. Accordingly, all the data, models, and programming code presented herein are provided under the FAIR \cite{wilkinson2016FAIR} (findable, accessible, interoperable, and reusable) conditions.  

The paper is structured as follows: Section \ref{sec:feat_inter} discusses the significant features that make the varying source training sets interoperable for satellite image classification, and Section \ref{sec:satimnet} introduces the {SatImNet} collection which organises existing training sets in an optimised and structural way. Section \ref{sec:fusion} demonstrates CNN models that have been trained on blended training sets and solve satisfactorily a satellite image classification and semantic segmentation task. Section \ref{sec:comp_platform} describes the computing platform over which the experimental process has been performed. Section \ref{sec:conclusion} underlines the contribution of the present work and outlines the way forward.

\section{Major Features of an Interoperable Training Set}
\label{sec:feat_inter}
In the following, we define the minimal and essential attributes that should characterise a training set when examined under the prism of interoperability and completeness. For best clarity, we have grouped the attributes according to three categories enumerated with their associated attributes~hereafter. 
 
\begin{enumerate}[leftmargin=*,labelsep=5mm]
\item[1.] \textit{Attributes related to the scope of the training data:}
\begin{enumerate}[leftmargin=1.5em,labelsep=4mm]
	\item[-] {Classification problem: denotes the family/genre of the supervised learning problem that the specific data set can serve in a more effective way;}
	\item[-] {Intended application: explains the primary purpose that the specific data set serves according to the data set designers;}
	\item[-] {Definition of classes: signifies how the class annotations are originally provided by the data set designer;}
	\item[-] {Annotation: provides information about the class label annotation, whether derived from a manual or automated procedure, or it is based on expert opinion, volunteering effort, or~organised crowd-sourcing; it serves as a qualitative indicator about the reliability of the provided data;}
	\item[-] {Verification: linked with the former feature, it refers as well to the belief level and reliability of the information transmitted by the data;}
	\item[-] {Licence: the existence of terms, agreements, and restrictions as were explicitly stated by the data publishers;}
	\item[-] {URL: the original web link to the data.}
	\end{enumerate}
	\end{enumerate}
	\begin{enumerate}[leftmargin=*,labelsep=5mm]
\item[2.] \textit{Attributes related to the usage and sustainability of the training data:}
\begin{enumerate}[leftmargin=1.5em,labelsep=4mm]
	\item[-] {Geographic coverage: reveals the terrain characteristics, the morphological features of the objects that shape the surface, and the landscape variability, as well as the potential irregularities covered by the candidate data set;}
	\item[-] {Timestamp: the image sensing time or the time window to which the image refers is crucial information for change detection and seasonality-based applications. This piece of information is closely related to the concept of temporal resolution but cannot be used interchangeably;}
	\item[-] {Data volume: helps to determine disk storage and memory requirements;}
	\item[-] {Data lineage: necessary information for precise reproducibility of the data processing workflow, including pre-processing transformations such as standardisation, normalisation, clipping of values, quantisation, projection, resampling, correction (atmospheric, terrain, cirrus), etc;}
	\item[-] {Name of classes: the semantic name that describes the category to which a single pixel, a~group of pixels, or a rectangular window (patch) belongs to;}
	\item[-] {Number of classes: shows the plurality and the exhaustiveness of the targets to be detected or~identified;}
	\item[-] {Naming convention: whether the file name conveys additional information such as sensing time, class name, location and so on;}
	\item[-] {Quality of documentation: a qualitative annotation assigned by the users (in this case, by us) about the existence of sufficient explanatory material;}
	\item[-] {Continuous development: a qualitative indicator about data sustainability, error correction and quality improvement that is based on the information provided by the data designers.}
	\end{enumerate}
	\end{enumerate}
	\begin{enumerate}[leftmargin=*,labelsep=5mm]
\item[3.] \textit{Intrinsic image attributes:}
\begin{enumerate}[leftmargin=1.5em,labelsep=4mm]
	\item[-] {File format: indicates file compression, the file reader and encoding/decoding type, availability of meta-information (e.g., GeoTIFF, PNG, etc.), number of channels/bands;}
	\item[-] {Image dimensions: a quick reference (image height and width) to estimate the batch size during the training phase, expressed as image rows $\times$ columns;}
	\item[-] {Number of bands: number of channels packed into a single file or number of separate single-band files belonging to a dedicated subfolder associated with the image name;}
	\item[-] {Name of bands: standard naming of the image channels like RGB (red/green/blue) or specific naming that follows the product convention such as the naming of Sentinel-2 products;}
	\item[-] {Data type per band: essential information about the range of band values that differentiates the data distributions and impacts the data normalisation/standardisation operations;}
	\item[-] {No data value: specifies the presence of invalid values that affect processes such as data normalisation and masking;}
	\item[-] {Spatial resolution: it determines what types of targets can be detected and indirectly points at the physical size of the training samples. Spatial resolution is often expressed in meters;}
	\item[-] {Spectral resolution: it refers to the capacity of a satellite sensor to measure specific wavelengths of the electromagnetic spectrum. In data fusion context, spectral resolution helps to compare and match image bands delivered by different sensors;}
	\item[-] {Temporal resolution: the amount of time that elapses before a satellite revisits a particular point on the Earth's surface. Although temporal resolution is an important attribute, there are no training sets that currently cover this aspect in detail;}
	\item[-] {Type of original imagery: a piece of information with reference to the sensor type and source, the availability of masks, the existence of geo-reference, and other auxiliary details;}
	\item[-] {Orientation: information referring mainly to image or target rotation/positioning; in the case of non-explicit statement, this feature contains basic photogrammetry information such as rectification and geometric correction;}
	\item[-] {Metadata: extra information that accompanies an image. It concerns mostly the geo-referenced and time-stamped images.}
\end{enumerate}
\end{enumerate}

Additional informative features could be the different/alternative usage of a data set with respect to other research works, number of citations, impact gauging, and the physical location (file path) at the local or remote storage system. In some cases, data publishers provide the date of the data set release, but this should not be confused with the formerly mentioned, pivotal attribute of timestamp.

For clarification purposes, we also detail the image classification applications that a training set with the above mentioned attributes could serve. Image classification is a generic term to describe the process of labelling an entire image or part of it. It differs from image segmentation wherein the image is partitioned in two or more connected sets of pixels; in this context, no semantics is needed since the segmentation can be based on pixel value similarities/differences and connectivity rules~\cite{soille2008}. Depending on the granularity of classification (pixel, group of pixels, block/window/patch), which~is strongly impacted by the spatial resolution of the input imagery (it can be equally extended to the temporal resolution as well), we distinguish the following applications that can be considered as a semantic segmentation (a label is assigned to every segment/part of the image) at different levels of~detail:
\begin{itemize}
\item Block/window/patch-based classification: In this case, the input image is divided in several overlapping or non-overlapping units (blocks/windows/patches) and all or part of these units receive a label by the classifier. The assumption under this configuration is that a considerable amount of pixels that compose the unit belong to one class, the one that has been assigned to the unit by the classifier. Labelling a unit signifies that the specific unit contains the target which is subject to detection. Target localisation is when drawing a rectangle which actually frames the image area that incorporates the target.
\item Group of pixels classification: A label is assigned to spatially connected pixels, forming patterns with concrete dimensions and structure that resemble pre-defined models of physical or artificial objects.  This type of application is known as instance segmentation or object delineation.
\item Pixel-wise classification: The target is every single pixel that can form a distinct class. The pixel labelling is performed by the classifier without considering the classification of the adjacent pixels. This type of operation occurs i) when the spatial resolution is lower than the real dimensions of the target (for instance, the finer resolution of the Sentinel-2 radiometric bands is 10 m, permissible for a building detection but not admissible for the detection of cars), or ii) for land cover classification or big areas identification such as the road or train network. 
\end{itemize}

The term target conveys the abstract concept of a pattern that is detectable and sometimes identifiable. It is subjected to the satellite sensor capacity according to the supported spatial, spectral, temporal, and radiometric resolution. Pixel targets are considered mostly for classification of land cover types such as forest, urban, water, crops, bare soil, etc., i.e., classes that may have irregular morphological characteristics such as shape or compactness. Group of pixels are considered for object detection or recognition. In this context, the pixel formation visually resembles real-world objects such as trees, houses, cars, boats, etc., usually targets with concrete structure, distinct boundaries, cohesion, and~other characteristics that make them separable from their surrounding area. Patches (rectangular windows) can be employed in both of the aforementioned applications. In object detection, the image patch encloses completely the target, i.e., comprises the group of pixels that form the object. In land cover classification, the rectangular window surrounds an ample number of pixels that represent a single land cover class or a mixture of land cover types. 

\section{{SatImNet} Collection}
\label{sec:satimnet}
In this section, we describe the initial edition of the {SatImNet} (Satellite Image Net) collection, a~compilation of seven open and free training sets targeting various EO applications. Then, we~elaborate on the rationale behind our choices for the data structuring. Lastly, we tabulate the training sets under consideration according to the defined attributes of Section \ref{sec:feat_inter}.

\subsection{Description of the Training Sets}
The initial edition of {SatImNet} consists of seven diverse training sets: 
\begin{enumerate}
\item {DOTA}: A large-scale Dataset for Object deTection in Aerial images, used to develop and evaluate object detectors in aerial images  \cite{Xia2018};
\item {xView}: contains proprietary images (DigitalGlobe's WorldView-3) from complex scenes around the world, annotated using bounding boxes \cite{lam2018};
\item Airbus-ship: combines Airbus proprietary data with highly trained analysts to support the maritime industry and monitoring services \cite{Airbus2018};
\item Clouds-s2-taiwan: contains Sentinel-2 True Colour Images (TCI) and corresponding cloud masks, covering the area of Taiwan \cite{Liu2019};
\item Inria Aerial Image Labeling: comprises aerial ortho-rectified colour imagery with a spatial resolution of 0.3 m and ground truth data for two semantic classes (building and no building)~\cite{maggiori2017};
\item {BigEarthNet-v1.0}: a large-scale Sentinel-2 benchmark archive consisting of Level-2A (L2A: Bottom Of Atmosphere reflectance images derived from the associated Level-1C products) Sentinel-2 image patches, annotated by the multiple land-cover classes that were provided from the CORINE Land Cover database of the year 2018 \cite{Sumbul2019};
\item {EuroSAT}: consists of numerous Level-1C (L1C: top-of-atmosphere reflectances in cartographic geometry) Sentinel-2 patches provided in two editions, one with 13 spectral bands and another one with the basic RGB bands; all the image patches refer to 10 classes and are used to address the challenge of land use and land cover classification \cite{helber2019}. 
\end{enumerate}

With regard to satellite imagery, one of the unique features is the spatial resolution which determines substantially the type of application. The high-resolution imagery provided by {DOTA}, {xView}, Airbus-ship, and Inria Aerial Image Labeling is suitable for object detection, localisation, and~semantic segmentation. The remaining three data sets are fitting mostly applications relevant to both image patch and pixel-wise classification.

As last note, we underline the fact that although the images of {xView} and {Airbus-ship} data sets have an ownership, they are provided for free under the licenses described in the subsequent Table \ref{tab:Dsets_attributes1}.

\subsection{{SatImNet} Data Model}
This section explains the rationale behind the chosen data model. The SatImNet collection has been structured in a modular fashion, preserving the unique characteristics of each constituent data set while providing a meta-layer that acts in a similar way as an ontology does, recording links and modelling relations among concepts and entities from the different data sets. The first two abstract layers of this meta-layer consist of the following keys: (1) built-up: residential, industrial, facilities, infrastructure, construction, areas; (2) transport means: vehicle, flying, vessel; (3) object: man-made; (4) natural areas: air, land, water. The third layer is composed by the leaf nodes (terminal nodes) representing all the classes of the seven data sets. While the intuitive choice was to consolidate the names of the classes, we decided to leave the original class names in order to retain a backward compatibility from {SatImNet} to the seven data sets. Accordingly, we preserve class names such as ``residential building'' although there is another leaf node ``bulding'' and a parent node ``residential''. Another consequence is that leaf nodes have duplicates, as it is the case of ``damaged building'' that belongs to more than one parent nodes such as the ``residential'' and ``industrial'' nodes.

\begin{figure}[H]
\centering
\includegraphics[width=0.98\textwidth]{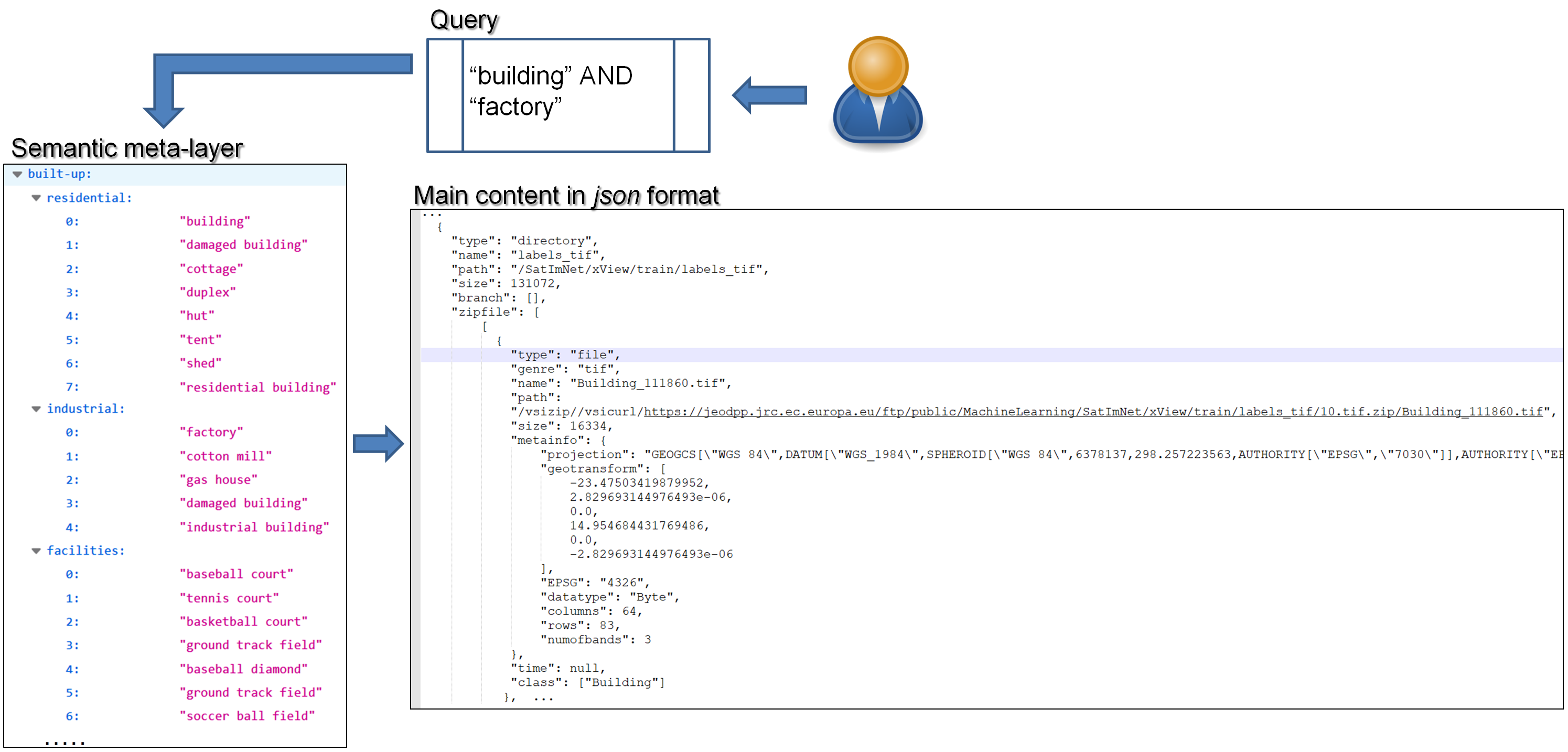}
\caption{Schematic representation of the information retrieval task: the user forms a query by selecting one or more keywords, e.g., \textit{building} and \textit{factory}. Then, the query passes first through the semantic meta-layer which seeks for conceptual similarities (e.g., \textit{built-up}, \textit{residential}, \textit{industrial}), and~then performs search across the \textit{json} files that retain information about the intrinsic attributes of every individual image.}
\label{fig:structure}
\end{figure}

Apart from the meta-layer that has formed as a lattice, the other structures that compose the data model are represented by nested short tree hierarchies which have been proven to be quite efficient in information retrieval tasks \cite{Baeza2011}. Accordingly, a \textit{json} file \textit{content\_public.json} has been created for each data set that contains mostly the intrinsic attributes of the images of the specific data set in a hierarchical, tree-based format. These attributes represent information such as the physical path of the file, its size in bytes, the file type (genre), the image acquisition time or the time the image refers to, the class label (if any), the meta-information like projection, number of bands and so on. Figure~\ref{fig:structure} displays a subset of the hierarchy and shows the typical route a query follows across the central semantic meta-layer and each information module which condenses the essential information that characterises every file.

Since the entire or part of the collection is accessible over the network, we selected a database-free solution for the tree hierarchies based on \textit{json} files. This portable layout grants a standalone character to the modules of the collection, independent of specialised software and transparent to the non-expert end-user. The lack of indexing which impacts critically the query speediness is tackled (whenever is feasible) by keeping the depth and breadth of every single tree in moderate sizes. A consequence of this is the creation of multiple \textit{json} files for each data set (e.g., 13 files in the case of {BigEarthNet-v1.0}). At this point, we underline the fact that the baseline system upon which we optimise all the processes is the {EOS} open-source distributed disk storage system developed at {CERN} \cite{EOS2015}, having as front-end a multi-core processing platform \cite{soille2017,Soille2018}. This configuration allows multi-tasking and is suitable for distributed information retrieval out of many files. One bounding condition set by the baseline system is the prevention of generating many small-sized files, given that {EOS} guarantees minimal file access latencies via the operation of in-memory metadata servers which replicate the entire file structure of the distributed storage system. For this reason, the files of the training sets have been zipped into larger archives, the size of which has been optimised in a way as to allow admissible information retrieval whilst sustaining efficient data transfer across the network. Reading individual files from \textit{zip} files can be achieved through various interfaces and drivers. In our case, we employ the open source Geospatial Data Abstraction Library (GDAL) \cite{gdal2020} which provides drivers for all the standard formats of raster and vector geospatial data. Jupyter notebooks demonstrating the execution of queries as well as the respective information retrieval from the \textit{json} files and subsequently by the \textit{zip} archives are referred to in Section \ref{sec:comp_platform}. Although the decision to zip the files was based on the technical characteristics of the multi-petabyte {EOS} open storage solution, the files can be of course unzipped if more suitable in other environments. To harmonise the class annotations provided in different file formats (json, geojson, text, and csv), binary or labelled image masks were created for every single training sample. Although the aforementioned data model has been optimised upon a distributed storage system, it turns out to be a general-purpose model built on the principles of simplicity, speediness, extensibility, customisation, and portability.

\subsection{Characterisation of the Data Sets}
The three attribute categories presented in Section \ref{sec:feat_inter} have been used as a basis to characterise the seven training sets of {SatImNet}. Tables \ref{tab:Dsets_attributes1}--\ref{tab:Dsets_attributes3} respectively show the results of this analysis. In Table \ref{tab:Dsets_attributes1}, the feature \textit{Conversion of class representation} has been added, indicating whether the original type of class annotations has been converted into an image mask upon creation of {SatImNet}. The dash ``-'' has been used to denote either non-existent information or non-clear definition.

\newpage
\paperwidth=\pdfpageheight
\paperheight=\pdfpagewidth
\pdfpageheight=\paperheight
\pdfpagewidth=\paperwidth
\newgeometry{layoutwidth=297mm,layoutheight=210 mm, left=2.7cm,right=2.7cm,top=1.8cm,bottom=1.5cm, includehead,includefoot}
\fancyheadoffset[LO,RE]{0cm}
\fancyheadoffset[RO,LE]{0cm}

\begin{table}[H]
 \centering
	\caption{Attributes related 
 to the scope of the training data.}
	    \label{tab:Dsets_attributes1}
    \scalebox{0.71}[0.71]{
      \begin{tabular}{llllllll}
        \toprule
        & \textbf{DOTA} & \textbf{xView} & \textbf{Airbus-Ship} & \textbf{Clouds-s2-} \textbf{Taiwan} & \textbf{Inria Aerial Image Labeling} & \textbf{BigEarthNet-} \textbf{v1.0} & \textbf{EuroSAT} \\
        \midrule
        \multirow{2}{*}{Classification problem} & \multirow{2}{*}{object detection} & \multirow{2}{*}{object detection} & \multirow{2}{*}{object detection} & \multirow{2}{*}{pixel-based detection} & pixel-based \&  & patch-based land  & patch-based land  \\
        &&&&&object detection&cover classification&cover classification\\
        \midrule
        \multirow{2}{*}{Intended application} & object detection in  & \multirow{2}{*}{\textsuperscript{(1)}} & \multirow{2}{*}{locate ships in images} & \multirow{2}{*}{clouds classification} & \multirow{2}{*}{building classification} & \multirow{2}{*}{land cover classification} & land use and land  \\
        &aerial images&&&&&&cover classification\\
        \midrule
        \multirow{2}{*}{Definition of classes} & \multirow{2}{*}{bounding boxes in txt} & bounding boxes  & \multirow{2}{*}{boxes encoding in csv} & \multirow{2}{*}{GeoTIFF images} & \multirow{2}{*}{GeoTIFF images} & \multirow{2}{*}{tags in json} & name of the files: RGB jpg; \\
        &&in geojson&&&&& 13-band GeoTIFF\\
        \midrule
        Conversion of class  & \multirow{2}{*}{txt to png} & \multirow{2}{*}{geojson to GeoTIFF} & \multirow{2}{*}{csv to png} & \multirow{2}{*}{no conversion }& \multirow{2}{*}{no conversion} & \multirow{2}{*}{no conversion} &\multirow{2}{*}{ no conversion} \\
        representation &&&&&&&\\
        \midrule
        Annotation & manual; experts & - & - & manual & \textsuperscript{(2)} & based on CLC 2018 & manual \\
        \midrule
        Verification & visual & - & - & - & visual & visual & visual \\
        \midrule
        \multirow{2}{*}{Licence} & \multirow{2}{*}{For academic purposes} & \multirow{2}{*}{CC BY-NC-SA 4.0} & \multirow{2}{*}{Non-commercial purposes} & - & Non-commercial  & \multirow{2}{*}{CDLA- Permissive-1.0} & - \\
        &&&&&purposes&&\\
        \midrule
        \multirow{2}{*}{URL} & \href{https://captain-whu.github.io/DOTA/dataset.html}{https://captain-whu.}  & \multirow{2}{*}{\url{http://xviewdataset.org}} & \href{https://www.kaggle.com/c/airbus-ship-detection/overview}{https://www.kaggle.com/c/} & \href{https://www.mdpi.com/2072-4292/11/2/119/s1}{https://www.mdpi.com/} & \href{https://project.inria.fr/aerialimagelabeling}{https://project.inria.fr/} & \multirow{2}{*}{\url{http://bigearth.net}} & \href{https://github.com/phelber/eurosat}{https://github.com/phelber/} \\
        &\href{https://captain-whu.github.io/DOTA/dataset.html}{github.io/DOTA/dataset.html}&&\href{https://www.kaggle.com/c/airbus-ship-detection/overview}{airbus-ship-detection/overview}&\href{https://www.mdpi.com/2072-4292/11/2/119/s1}{2072-4292/11/2/119/s1}&\href{https://project.inria.fr/aerialimagelabeling}{aerialimagelabeling}&&\href{https://github.com/phelber/eurosat}{eurosat}\\
        \bottomrule 
      \end{tabular}}
      \begin{tabular}{cccccccc}
\multicolumn{1}{c}{\footnotesize \textsuperscript{1} Enables discovery of more object classes; improves detection of fine-grained classes. \textsuperscript{2} Combines public domain imagery with public domain official building footprints.}
 \end{tabular}
\end{table}
\unskip
\begin{table}[H]
 \centering
    \caption{Attributes related to the usage and sustainability of the training data.}
     \label{tab:Dsets_attributes2}
       \scalebox{0.95}[0.95]{\begin{tabular}{llllllll}
        \toprule 
        & \textbf{DOTA} & \textbf{xView} & \textbf{Airbus-Ship} & \textbf{Clouds-s2-} \textbf{Taiwan} & \textbf{Inria Aerial Image Labeling} & \textbf{BigEarthNet-} \textbf{v1.0} & \textbf{EuroSAT} \\
        \toprule 
        Geographic coverage & variable & 1415 km$^2$  \textsuperscript{(1)} & variable & Taiwan & 810 km$^2$ \textsuperscript{(2)} & 10 European countries \textsuperscript{(3)} & 34 European cities  \\ 
        \midrule
        Timestamp & - & - & - & May 2018 & - & June 2017--May 2018 & variable \\
        \midrule
        Data volume & 19.9 GB & 36.4 GB & 29.5 GB & 123 MB & 25.3 GB & 106 GB & 2.88 GB \\ 
        \midrule
        Data lineage & - & - & - & - & - & sen2cor & L1C \\
        \midrule
        Name of classes & \textsuperscript{(4)} & \textsuperscript{(5)} & ship/no ship & cloud/no cloud & building/no building & CLC nomenclature \textsuperscript{(6)} & \textsuperscript{(7)} \\
        \midrule
        Number of classes & 15 & 60 & 2 & 2 & 2 & 12 \textsuperscript{(8)} & 10 \\
        \midrule
        Naming convention & no & no & no & yes & yes & yes & at class level \\ 
        \midrule
        Quality of documentation & good & moderate & not detailed & good & good & very good \textsuperscript{(9)} & good \\ 
        \midrule
        Continuous development & - & - & no & no & no & yes & - \\
        \bottomrule 
      \end{tabular}}
     \begin{tabular}{cccccccc}
\multicolumn{1}{p{\textwidth -.88in}}{\footnotesize \textsuperscript{1} part of cities around the world. \textsuperscript{2} towns around the world: Austin, Chicago, Kitsap County, Western Tyrol, Vienna. \textsuperscript{3} Austria, Belgium, Finland, Ireland, Kosovo, Lithuania, Luxembourg, Portugal, Serbia, Switzerland. \textsuperscript{4} plane, ship, storage tank, baseball diamond, tennis court, basketball court, ground track field, harbor, bridge, large vehicle, small vehicle, helicopter, roundabout, soccer ball field, swimming pool (and one additional: container-crane). \textsuperscript{5} various aircraft types, vehicles, boats/vessels, buildings, and man-made objects such as containers, pylons, and towers. \textsuperscript{6} level-3 CORINE Land Cover class labels. \textsuperscript{7} industrial buildings, residential buildings, annual crop, permanent crop, river, sea \& lake, herbaceous vegetation, highway, pasture, forest. \textsuperscript{8} multiple classes per patch. \textsuperscript{9} \url{http://bigearth.net/static/documents/BigEarthNetManual.pdf}.}
\end{tabular}
   
\end{table}
\newpage
\restoregeometry
\paperwidth=\pdfpageheight
\paperheight=\pdfpagewidth
\pdfpageheight=\paperheight
\pdfpagewidth=\paperwidth
\headwidth=\textwidth

\newpage
\paperwidth=\pdfpageheight
\paperheight=\pdfpagewidth
\pdfpageheight=\paperheight
\pdfpagewidth=\paperwidth
\newgeometry{layoutwidth=297mm,layoutheight=210 mm, left=2.7cm,right=2.7cm,top=1.8cm,bottom=1.5cm, includehead,includefoot}
\fancyheadoffset[LO,RE]{0cm}
\fancyheadoffset[RO,LE]{0cm}

\begin{table}[H]
   \centering
    \caption{Intrinsic image attributes.}
    \label{tab:Dsets_attributes3}
      \scalebox{0.85}[0.85]{ \begin{tabular}{llllllll}
        \toprule 
        & \textbf{DOTA} & \textbf{xView} & \textbf{Airbus-Ship} & \textbf{Clouds-s2-} \textbf{Taiwan} & \textbf{Inria Aerial Image Labeling} & \textbf{BigEarthNet-} \textbf{v1.0} & \textbf{EuroSAT} \\
        \toprule 
		\multirow{2}{*}{File format} & \multirow{2}{*}{png} & \multirow{2}{*}{GeoTIFF} & \multirow{2}{*}{jpg} & \multirow{2}{*}{GeoTIFF} & \multirow{2}{*}{GeoTIFF} & \multirow{2}{*}{GeoTIFF} & RGB: jpg;  \\ 
		&&&&&&&13-band: GeoTIFF\\
        \midrule        
        Image dimensions & from $800\times800$ to  & \multirow{2}{*}{various small patches} & \multirow{2}{*}{$768\times768$} & \multirow{2}{*}{$224\times224$} & \multirow{2}{*}{$5000\times5000$} & $120\times120$, $60\times60$,  & \multirow{2}{*}{$64\times64$ }\\ 
         (rows $\times$ cols)&about $4000\times4000$&&&&&or $20\times20$&\\
        \midrule
        Number of bands & 3 & 3 & 3 & 10 & 3 & 12 & 13-band \& RGB \\
        \midrule
        Name of bands & RGB & RGB and 8-band & RGB & Sentinel-2 bands & RGB & Sentinel-2 L2A bands & class name \\ 
        \midrule
        Data type per band & 8 & 8 & 8 & 16 & 8 & 16 & RGB: 8; 13 band: 16 \\
        \midrule
        No data value & - & 0 & - & - & 0 \textsuperscript{(2)} & - & - \\ 
        \midrule
        Spatial resolution & variable & 0.3 m & - & 20 m & 0.3 m & 10 m; 20 m; 60 m & 10 m; 20 m; 60 m \\ 
        \midrule
        Spectral resolution & - & - & - & \textsuperscript{(1)} & - & \textsuperscript{(1)} & \textsuperscript{(1)} \\ 
        \midrule
        \multirow{2}{*}{Type of original imagery} & multiple sensors;  & \multirow{2}{*}{WV-3; geo-referenced} & \multirow{2}{*}{non geo-referenced} & \multirow{2}{*}{geo-referenced} & \multirow{2}{*}{geo-referenced} & Sentinel-2 patches;  & Sentinel-2 patches;  \\ 
        &non geo-referenced&&&&&geo-referenced&geo-referenced\\
        \midrule
        Orientation & variable & ortho-images & ortho-images & S2 L1C ortho-images & ortho-images & S2 L2A ortho-images & S2 L1C ortho-images \\
        \midrule
        Metadata & no & yes & no & yes & yes & yes & yes \\
        \bottomrule 
      \end{tabular}}
      \begin{tabular}{cccccccc}
\multicolumn{1}{c}{\footnotesize \textsuperscript{1} \url{https://sentinel.esa.int/web/sentinel/missions/sentinel-2/instrument-payload/resolution-and-swath}.  \textsuperscript{2} explicitly defined in some cases only.}
\end{tabular}
\end{table}

\newpage
\restoregeometry
\paperwidth=\pdfpageheight
\paperheight=\pdfpagewidth
\pdfpageheight=\paperheight
\pdfpagewidth=\paperwidth
\headwidth=\textwidth

\section{Fusion of Heterogeneous Training Sets: Experimental Results}
\label{sec:fusion}
This section shows the added value of {SatImNet} for enhanced information extraction from Earth Observation products. The fusion of training sets from multiple sources is challenging for the data are created with different technical specifications: source imagery, pre-processing workflow, assumptions, manual/automatic refinement, different spatial and temporal resolution, etc. Within {SatImNet} context, we investigate data fusion by conducting experiments demonstrating a satellite image classification and segmentation application based on CNN models. Working within the open and free data framework, we decided to employ satellite imagery provided by the Copernicus Sentinel-2 (S2) \cite{Sentinel2_2018} mission as input to the models since the S2 products are delivered for free and the experiments can be reproduced by anyone. The highest spatial resolution of S2 products is 10 m and fits better with applications such as land cover classification. This condition guided us to select the appropriate training data and exclude data sets that match better to object detection in very high resolution imagery. We note here that the objective of the {SatImNet} collection is not to provide a single big data set for any type of image classification or segmentation application but rather to offer varying source data that are homogenised to the greatest extent possible, thereby facilitating the user to select and combine different data sets that fit her purpose.

Two examples of applications are illustrated in this section. The first one deals with land cover classification formulated as a patch-based classification problem. From the {SatImNet} collection, we~chose two sets: (i) the {EuroSAT}, and (ii) the {BigEarthNet-v1.0} (see Tables \ref{tab:Dsets_attributes1}--\ref{tab:Dsets_attributes3} for the technical characteristics). We underline that the {EuroSAT} L1C data set has been used as a backbone set (main~corpus). In some of the experiments, this set has been enriched with training samples selected from the {BigEarthNet-v1.0} L2A data set: (i) we added 1000~samples from the {BigEarthNet-v1.0} \textit{Annual crops associated with permanent crops} class to the {EuroSAT} class \textit{annual crop}, and (ii) we augmented the size of the {EuroSAT} class \textit{forest} with 1000~samples taken randomly from the {BigEarthNet-v1.0} classes \textit{broad-leaved forest}, \textit{coniferous~forest}, and \textit{mixed forest}.

The second application is a semantic segmentation problem. In addition to the two above mentioned data sets, we chose the following two products:
\begin{enumerate}
\item In connection with the water class, we used the Global Surface Water (GSW) \cite{Pekel2016}, a collection of global and spatio-temporal water maps created from individual full-resolution 185 km$^2$ global reference system II scenes (images) acquired by the Landsat 5, 7, and 8 satellites and distributed by the United States Geological Survey (USGS). Two out of the ten {EuroSAT} classes refer to water variants (\textit{river} and \textit{sea lake}), depicting areas that are partially or totally covered by water. From~the {BigEarthNet-v1.0}, we randomly selected image patches referring to the classes \textit{coastal lagoons} and \textit{sea and ocean} (1000 samples from each category).
\item With regard to the {EuroSAT} classes \textit{industrial} and \textit{residential}, the image segmentation was based on the European Settlement Map \cite{corbane2019} (ESM 2015, R2019 dataset), a spatial raster data set that is mapping human settlements in Europe based on Copernicus Very High Resolution (VHR) optical coverage, having 2015 as the reference year. From the {BigEarthNet-v1.0}, we randomly selected image patches referring to the classes \textit{continuous urban fabric} and \textit{discontinuous urban fabric} (500~samples from each category). 
\end{enumerate}

The specific data fusion approach is just one of the many combinations and associations someone could follow, and can be deemed as a representative scheme rather than the optimal strategy. 

The selected {BigEarthNet-v1.0} image patches have been resized from their original size of \mbox{$120\times120$} to $64\times64$ images by applying a Gaussian filter to smooth the data and then by using bi-linear interpolation. On the basis of both {EuroSAT} and {BigEarthNet-v1.0} geo-referenced image patches, we~warped and clipped the GSW 2018 yearly classification layer, producing in that way the necessary water masks. Similarly, we clipped the 10 m up-scaled ESM and considered all the pixels pointing at residential and non-residential built-up areas. The non-residential built-up areas refer to detected industrial buildings, big commercial stores, and facilities. All the produced masks were resampled to 10 m spatial resolution using nearest neighbour interpolation. These masks compose an auxiliary data set that has been added to the {SatImNet} collection under the name {BDA}.

\subsection{Convolutional Neural Network Modelling}
Although CNNs have been experimentally proved to be more adequate for object \mbox{detection \cite{Cheng2016,Li2018}} and semantic segmentation \cite{Witharana2014, Audebert2017} in very high spatial resolution ($\leq$5 m), there is lately a considerable number of works \cite{SHENDRYK2019,SHARMA2017,Syrris2019,LiuSh2019} demonstrating promising results at coarser spatial resolutions such as those of the S2 imagery. Nevertheless, the majority of these works focus on image patch and not on pixel-wise classification, which is a more complex problem.

\begin{figure}[H]
\centering
\includegraphics[scale=0.40]{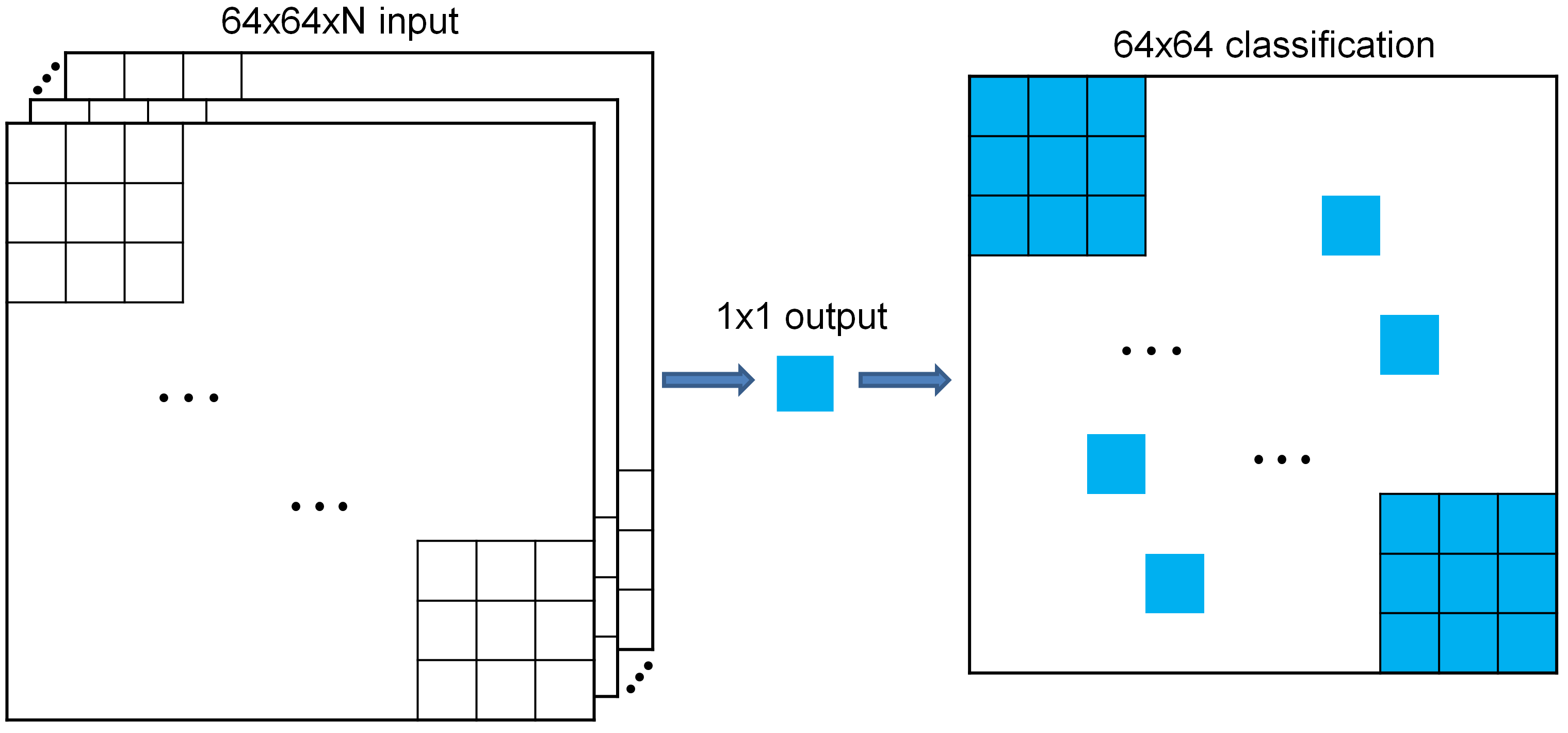}\\
\includegraphics[scale=0.40]{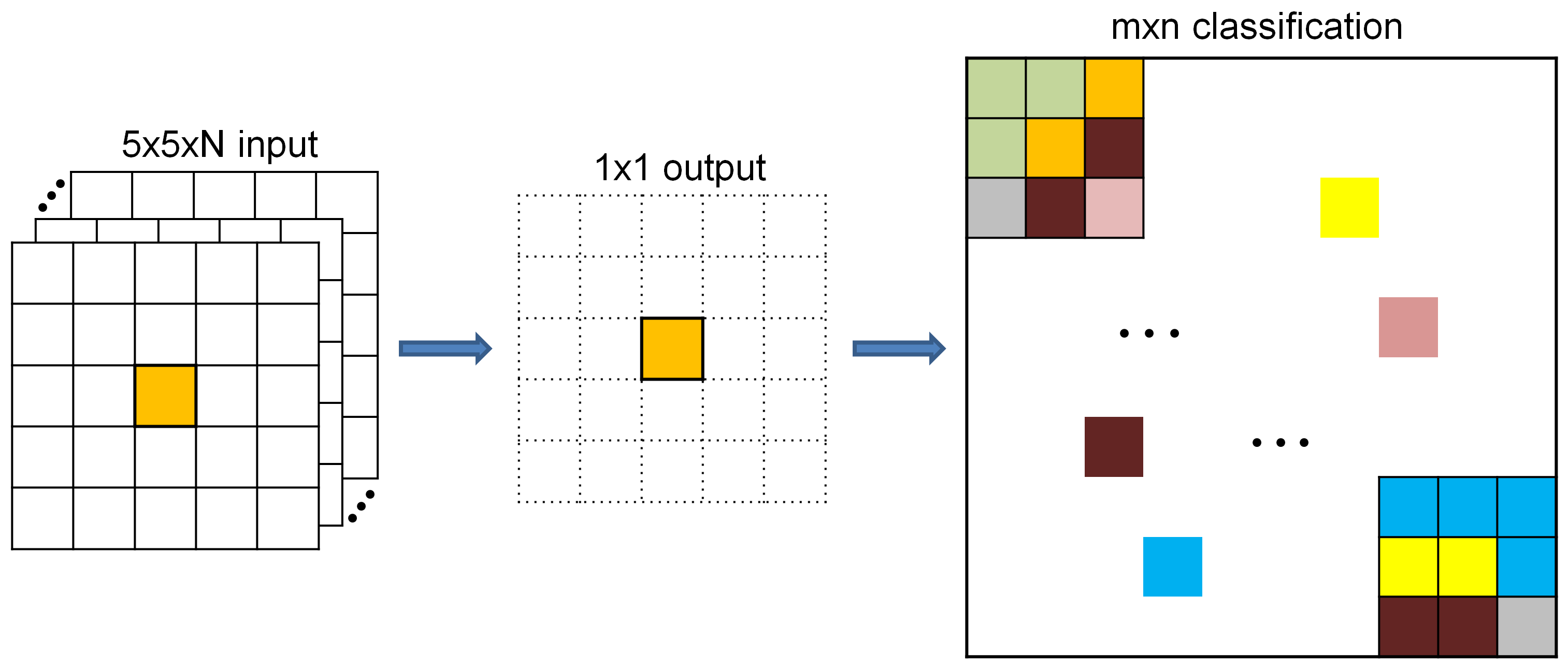}
\caption{Input and output with respect to the CNN-class model (\textbf{top}) and both the CNN-segm and CNN-dual models (\textbf{bottom}). The variable $N$ signifies the number of bands. The variables $m$ and $n$ denote the number of rows and columns, respectively, of the output image.}
\label{fig:kernels}
\end{figure}

The experimental configuration in this study has been chosen to explore the challenging problem of image segmentation together with image classification in the spatial resolution of 10 m of the S2 products. In order to assess better the added value of the data fusion, rather than using pre-trained models, we have designed and tested two customised lightweight CNN approaches instead, trained from scratch for 100 epochs. The input--output schema for the CNN models is depicted in Figure \ref{fig:kernels}.

\begin{figure}[H]
\centering
\includegraphics[width=0.75\textwidth]{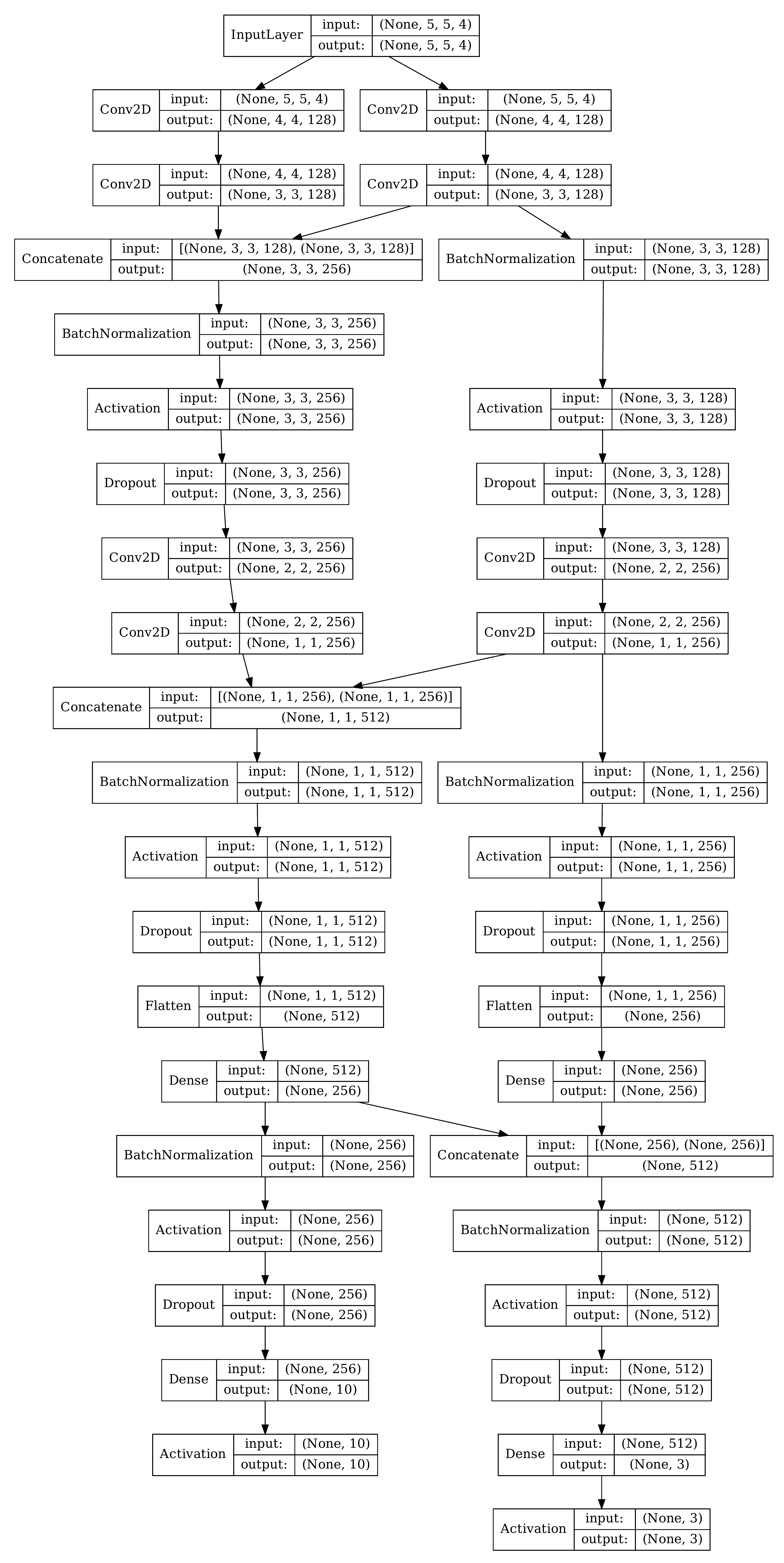}
\caption{A two-branch dual output CNN topology for image segmentation.}
\label{fig:topo1}
\end{figure}

The first CNN architecture (named CNN-dual) is a two-branch dual output CNN architecture that segments the image according to two classification schemas: \textit{left-branch output} is the classification result of assigning to each pixel one of the 10 {EuroSAT} classes (annual crop, forest, herbaceous vegetation, highway, industrial, pasture, permanent crop, residential, river, and sea lake), and \textit{right-branch output} is the pixel-wise classification result with reference to the aggregation classes water, built-up and other, as~instructed by the {GSW} and {ESM} layers. The input to such a model is a $5$ rows $\times$ $5$ columns $\times$ \textit{N} bands image (Figure \ref{fig:kernels}). Table \ref{tab:CNNparams} summarises the basic parametrisation of the model. The same classification task could be formulated as a 12-class problem modelled by a single branch CNN; nevertheless, experimental results showed that CNN-dual provides consistently better results. There~are three layer-couplings which intertwine the intermediate outputs across the two branches and sufficiently high structural capacity. This neural network architecture should not be confused with the twin neural network topology (Siamese) that uses the same weights while pairing two different inputs to compute comparable outputs. Figure \ref{fig:topo1} displays the CNN-dual model for the segmentation of S2 image patches.

\begin{figure}[H]
\centering
\includegraphics[scale=0.185]{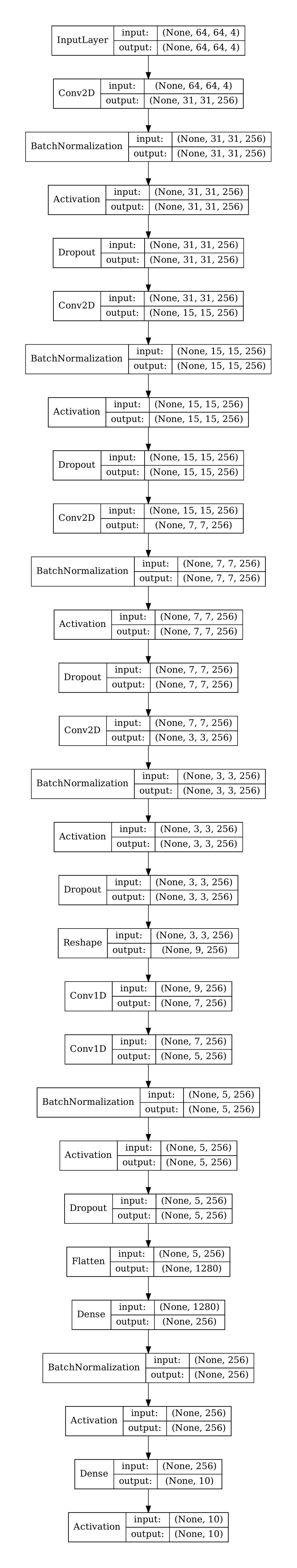}
\includegraphics[scale=0.2]{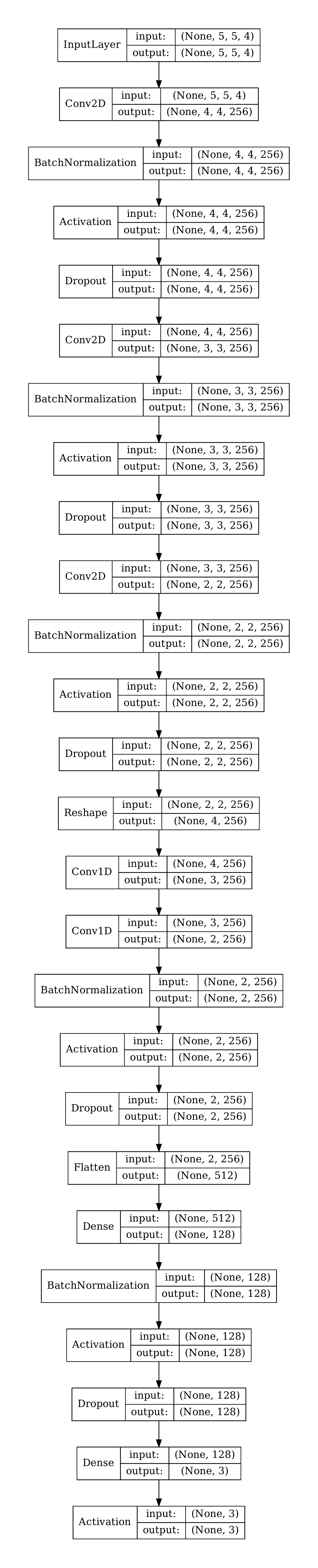}
\caption{Two independent CNN topologies: the left one performs a patch-based classification and the right one a pixel-wise classification.}
\label{fig:topo2topo3}
\end{figure}

The second CNN approach comprises two independent networks. The left network (CNN-class) as it appears in Figure \ref{fig:topo2topo3} takes a $64$ rows $\times$ $64$ columns $\times$ \textit{N} bands image and performs patch-based classification, i.e., it assigns one label from the 10 {EuroSAT} classes to all the $64\times64$ image pixels (Figure~\ref{fig:kernels}). The right network (CNN-segm) has been designed for image segmentation according to the three aggregated classes (water, built-up, and other). CNN-segm is applied solely on the $64\times 64\times N$ image patches classified by CNN-class as \textit{water} or \textit{built-up}. In this case, the $64\times 64\times N$ array disintegrates in blocks of size $5\times 5\times N$ following a sliding-window approach. Table \ref{tab:CNNparams} shows the choices for the basic parameters of the two models.

\begin{table}[H]
\centering
\caption{Parametrisation of the customised CNN models.}
\label{tab:CNNparams}
\scalebox{0.65}[0.65]{\begin{tabular}{ccccccccc}
\toprule
\multirow{2}{*}{\textbf{ Model}} & \textbf{\# Trainable} & \textbf{Activation} & \textbf{Dropout} & \textbf{Random Weights} & \textbf{Batch } & \textbf{Loss } & \multirow{2}{*}{\textbf{Optimiser}} & \multirow{2}{*}{\textbf{Output}} \\
& \textbf{Parameters}& \textbf{Function}& \textbf{Rate}&\textbf{Initial- Isation}&\textbf{Normal- Isation}&\textbf{Function}&&\\
\midrule
\multirow{2}{*}{CNN-dual }& \multirow{2}{*}{1,259,277} & relu   & \multirow{2}{*}{0.1} & He uniform   & \multirow{2}{*}{\checkmark \cite{ioffe2015}} & categorical  & stochastic gradient   & 10 classes  \\
&&(last layer: softmax)&&variance scaling \cite{He2015}&&cross- entropy&descent (0.01 learning rate)&and 3 classes\\
\midrule
\multirow{2}{*}{CNN-class} & \multirow{2}{*}{2,507,018} & relu  & \multirow{2}{*}{0.09} & Xavier normal  & \multirow{2}{*}{\checkmark} & categorical  & Adam \cite{Kingma2014}  & \multirow{2}{*}{10 classes} \\
&&(last layer: softmax) &&initializer \cite{Glorot2010}&&cross- entropy&(0.01 learning rate)&\\
\midrule
\multirow{2}{*}{CNN-segm} & \multirow{2}{*}{860,163} & tanh & \multirow{2}{*}{0.1} & Xavier  & \multirow{2}{*}{\checkmark} & categorical  & Adam  & \multirow{2}{*}{3 classes} \\
&&(last layer: softmax) &&normal initializer&&cross- entropy&(0.001 learning rate)&\\
\bottomrule
\end{tabular}}
\end{table}
\unskip

\begin{figure}[H]
\centering
\includegraphics[scale=0.55]{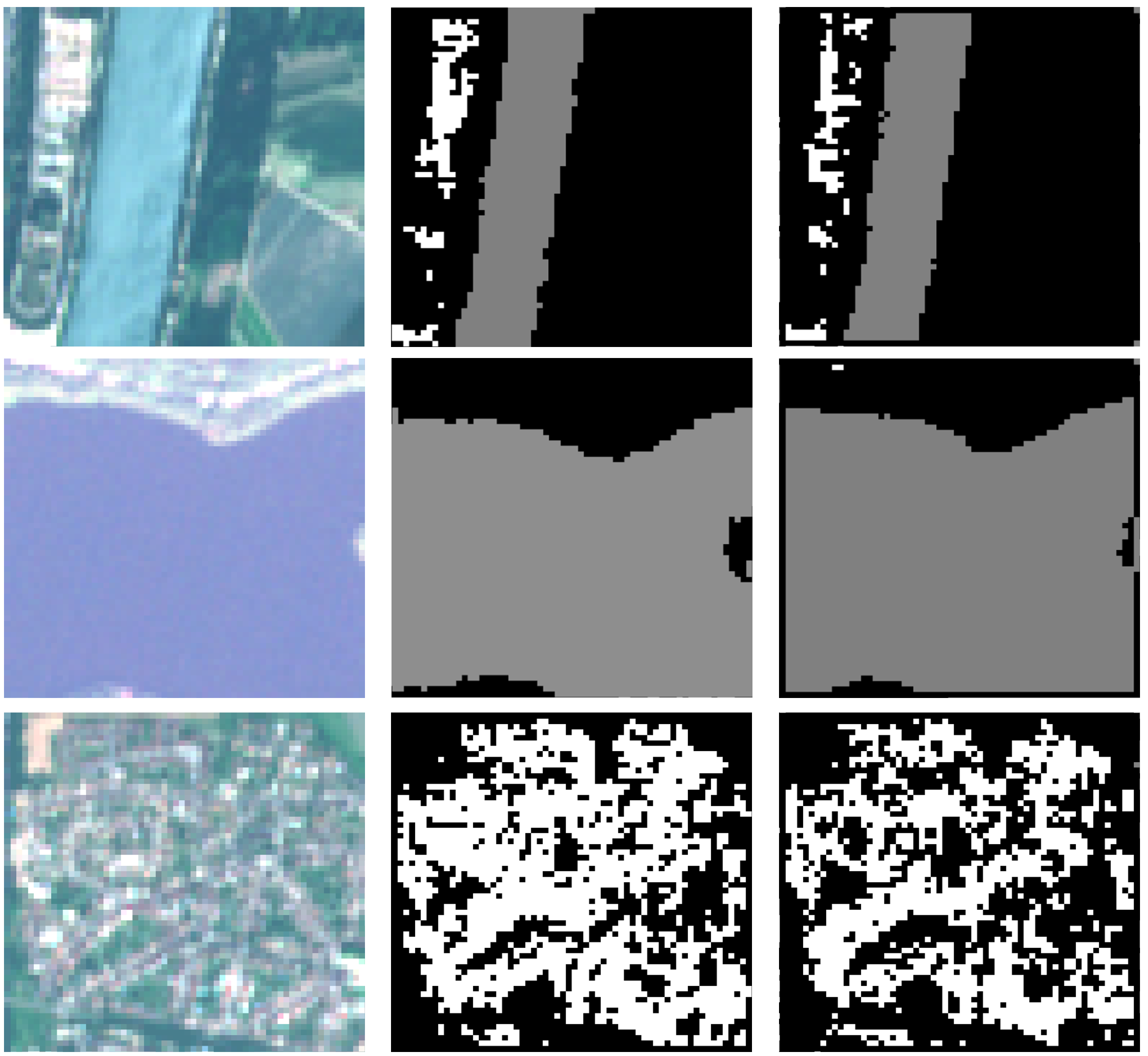}
\caption{Image patches ($64\times64$) of the {EuroSAT} data set kept out for testing. The first column displays the RGB composition, the middle column shows the output of the CNN-segm, and the last column displays the segmentation result of the CNN-dual model. The CNN-class classifies correctly the three image patches to \textit{river}, \textit{sea lake}, and \textit{residential}, respectively.}
\label{fig:snapshots}
\end{figure}

All the above mentioned parametrisations as well as the proposed CNN topologies derived from an extensive repetitive experimental process (threefold grid search). We note here that the purpose of this case study is not to conduct a comparison analysis of widely accepted CNN-based classification or segmentation models against the proposed ones. The presented CNN topologies are lightweight modelling approaches useful for evaluating the impact an enriched training set brings on the classification performance. Table \ref{tab:res_accuracy} depicts the classification accuracy in terms of two metrics, the F1-score and the Kappa score, while considering the four S2 bands with 10~m spatial resolution. Regarding the CNN-dual results, there are two values for each metric corresponding to 10 (left) and 3 (right) classes, respectively. Figure \ref{fig:snapshots} displays some indicative screenshots.

\unskip

We concluded also that the four S2 10 m bands, blue (B02), green (B03), red (B04), and NIR (B08) give for the most part of the performed experiments the most consistent results. Actually, we found out that there is an intense dissimilarity between the data distributions of the {EuroSAT}-provided bands B09 and B10 (the data set creators claim that all the S2 images have been downloaded via Amazon S3) and the respective bands of the MSI Sentinel-2 products we downloaded from the Copernicus Open Access Hub (\url{https://scihub.copernicus.eu/}).

\begin{table}[H]
\centering
\caption{Accuracy performance by the proposed CNN modelling approaches, computed via tenfold cross-validation. The training, validation, and testing sets have been partitioned according to the rule 80/10/10. The subheadings 10cl and 3cl refer to 10-class and 3-class result, respectively.}
\label{tab:res_accuracy}
\scalebox{0.95}[0.95]{\begin{tabular}{lllcccc}
\toprule
\textbf{ Model }& \textbf{Training Set} & \textbf{Testing Set} & \multicolumn{2}{c}{\textbf{F1-Score (\%)}} & \multicolumn{2}{c}{\textbf{Kappa Score (\%)}} \\
\midrule
\multicolumn{3}{l}{\textbf{Patch-based classification}} & \multicolumn{2}{c}{10cl} & \multicolumn{2}{c}{10cl} \\
\cmidrule(lr){4-5} \cmidrule(lr){1-7}
CNN-class & {EuroSAT} & {EuroSAT} & \multicolumn{2}{c}{96.65} & \multicolumn{2}{c}{96.28} \\
CNN-class & {EuroSAT} & {EuroSAT} \& {BigEarthNet-v1.0} & \multicolumn{2}{c}{87.23} & \multicolumn{2}{c}{84.84} \\
CNN-class & {EuroSAT} \& {BigEarthNet-v1.0} & {EuroSAT} & \multicolumn{2}{c}{98.76} & \multicolumn{2}{c}{98.62} \\
CNN-class & {EuroSAT} \& {BigEarthNet-v1.0} & {EuroSAT} \& {BigEarthNet-v1.0} & \multicolumn{2}{c}{94.44} & \multicolumn{2}{c}{93.77} \\
\cmidrule(lr){4-5} \cmidrule(lr){1-7}
\multicolumn{3}{l}{\textbf{Image segmentation}} & 10cl & 3cl & 10cl & 3cl \\ 
\cmidrule(lr){4-5} \cmidrule(lr){1-7}
CNN-dual & {EuroSAT} & {EuroSAT} & 72.13 & 83.70 & 70.71 & 66.51 \\
CNN-dual & {EuroSAT} & {EuroSAT} \& {BigEarthNet-v1.0} & 67.33 & 77.17 & 66.12 & 61.71\\
CNN-dual & {EuroSAT} \& {BigEarthNet-v1.0} & {EuroSAT} & 77.29 & 88.01 & 75.52 & 70.99 \\
CNN-dual & {EuroSAT} \& {BigEarthNet-v1.0} & {EuroSAT} \& {BigEarthNet-v1.0} & 74.89 & 86.89 & 72.15 & 67.98\\
\bottomrule
\end{tabular}}
\end{table}
 \unskip

\begin{figure}[H]
\centering
\includegraphics[scale=0.55]{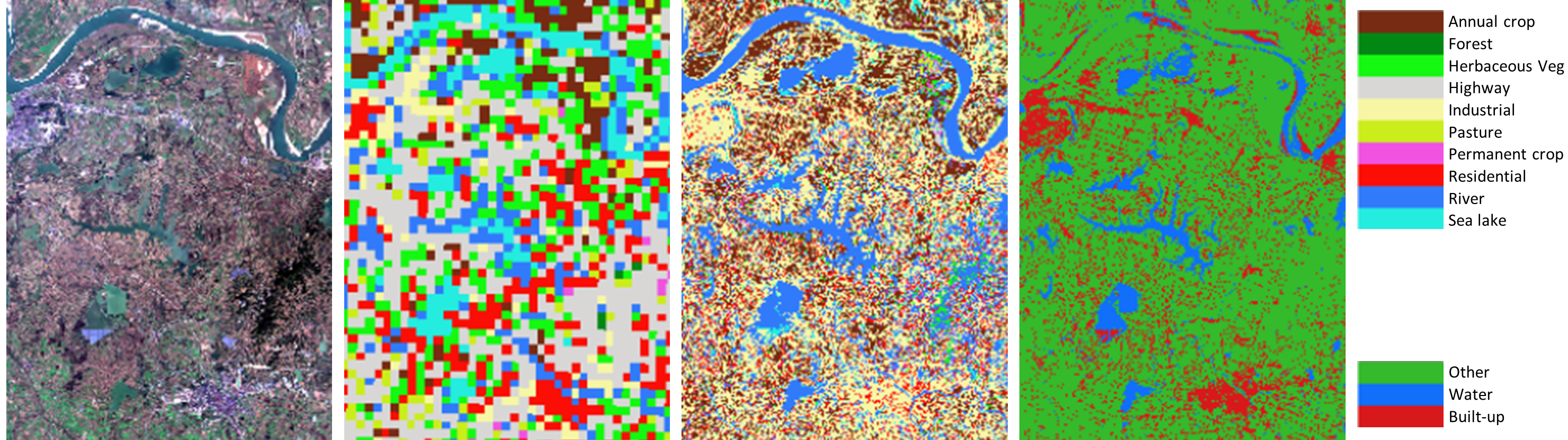} \\
\includegraphics[scale=0.55]{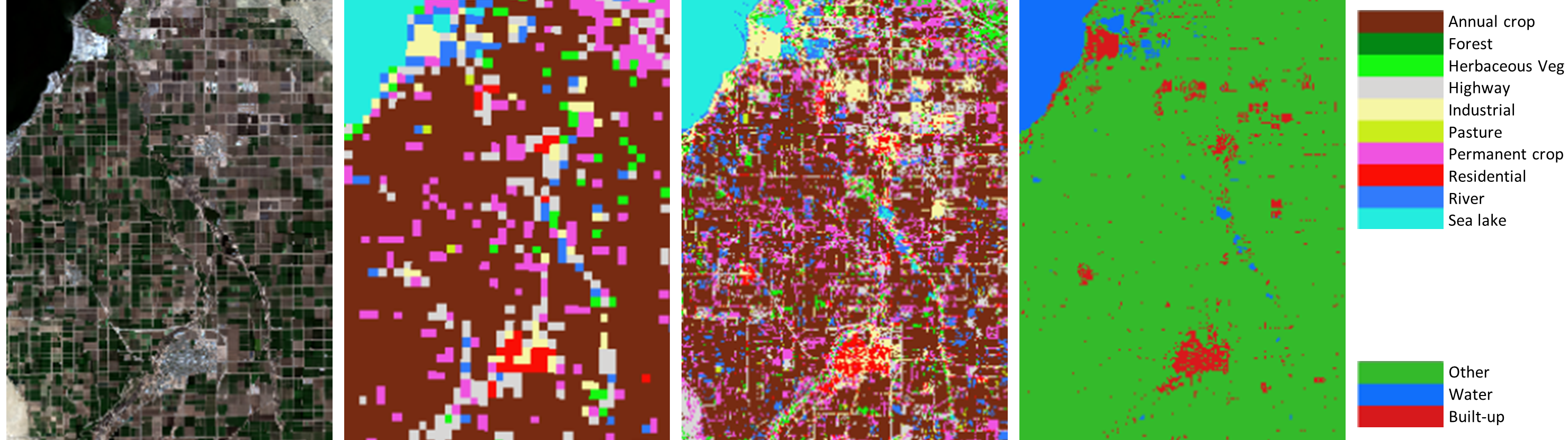}
\caption{Transfer learning on geographic locations different than the location from which the training samples have been selected. Two exemplar areas from China and USA in the first and second row, respectively; all the training samples have been selected from Europe. The columns from left to right: (i)~RGB, (ii) CNN-class, (iii) 10-class CNN-dual, and (iv) 3-class CNN-dual output.}
\label{fig:transferlearning}
\end{figure}

\begin{figure}[H]
\centering
\includegraphics[scale=0.55]{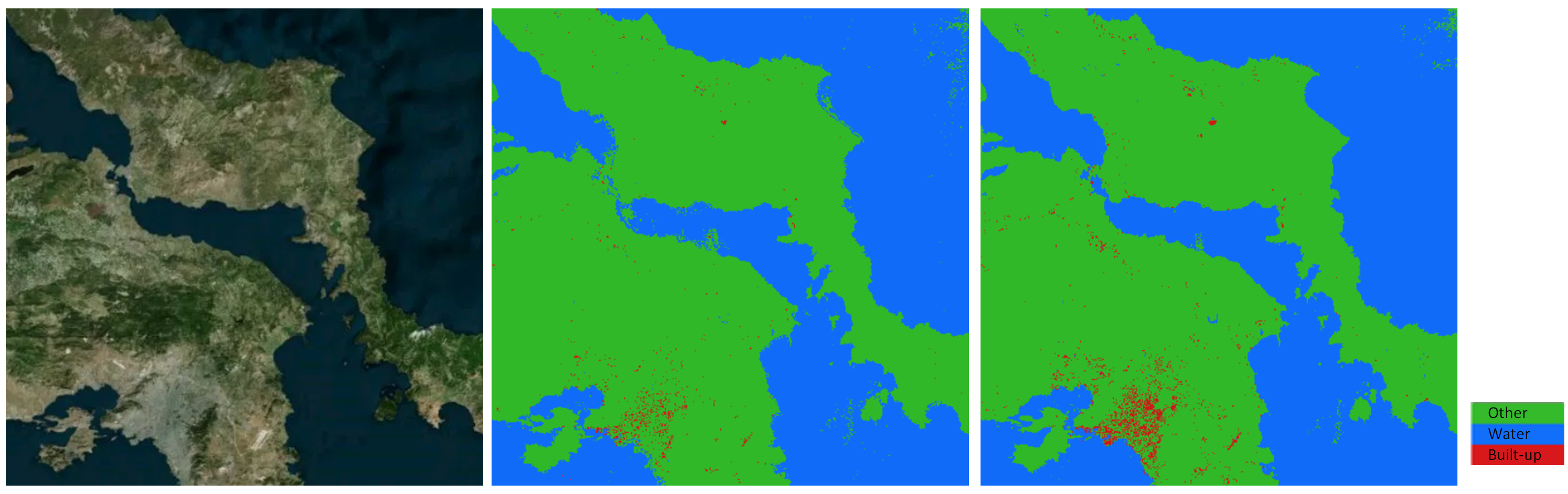}
\caption{Testing model robustness on L2A product (geographic area in Europe). The columns from left to right: (i)~RGB, (ii) 3-class CNN-segm, and (iii) 3-class CNN-dual output.}
\label{fig:L2A_transferlearning}
\end{figure}

\begin{figure}[H]
\centering
\includegraphics[scale=0.33]{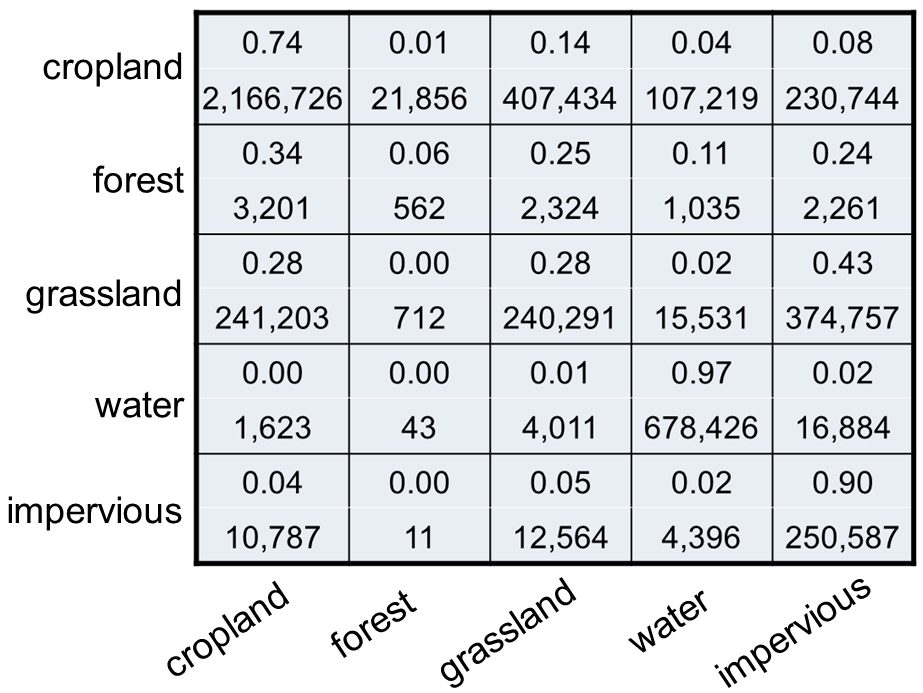}
\caption{Confusion matrix between the CNN-dual segmentation output and the {FROM-GLC} land cover map at 30 m spatial resolution. The top row in each cell represents the agreement percentage and the bottom row the actual number of classified pixels.}
\label{fig:confmatrix}
\end{figure}

Visual inspection of the classification results over (i) additional S2 MSI images (i.e., images not included in the collected data sets), (ii) geographic areas outside of the European continent where the training patches have been extracted, such as in China and USA (Figure \ref{fig:transferlearning}), and (iii) Level-2A (surface reflectance in cartographic geometry) S2 products (Figure \ref{fig:L2A_transferlearning}), while the majority of the training samples have been derived by L1C images, is in accordance with the results obtained from the L1C tests. The~visual inspection confirmed an agreement of more than 60\% with respect to the segmentation results and around 80\% for the patch-based classification. There is a consistent confusion among the classes \textit{residential}, \textit{industrial}, and \textit{highway}, as well as among the classes \textit{forest}, \textit{herbaceous vegetation}, and~\textit{pasture}. The~latter confusion is attributed mostly on the seasonality whereas the former happens as a result of the similar radiometric signature of the build-up structures. We observed also a known problem with the misclassification of the shadow cast by mountains or forest trees to one of the water classes. This can be potentially solved by incorporating training samples which represent this pattern. To quantitatively affirm our findings derived from visual observation, we have included an indicative confusion matrix (Figure \ref{fig:confmatrix}) that displays the agreement between the classification result of the model CNN-dual when applied over an S2 product covering an area in USA (Brawley---south region of Salton sea) and the global land cover product {FROM-GLC} \cite{Gong2019}. We employ the term agreement and not accuracy since the class nomenclature differs significantly between the two layers. To bridge the two layers, we applied the re-classification schema of Table \ref{tab:reclass}. The {FROM-GLC} classes \textit{tundra}, \textit{bareland}, and \textit{snow/ice} have not been considered. The comparison has been performed at the spatial resolution of {FROM-GLC} (30 meters) by up-scaling the segmentation output through the statistical \textit{mode} operator.

\unskip
\unskip
\unskip
\begin{table}[H]
\centering
\caption{Re-classification of CNN-dual segmentation output according to {FROM-GLC} nomenclature.}
\label{tab:reclass}
\begin{tabular}{ccc}
\toprule
 \textbf{{EuroSAT} Class} & & \textbf{{FROM-GLC} Class} \\
\midrule
annual crop, permanent crop & ~~~$\rightarrow$~~~ & cropland \\
forest & ~~~$\rightarrow$~~~ & forest \\
herbaceous vegetation, pasture & ~~~$\rightarrow$~~~ & grassland, shrubland \\
highway, industrial, residential & ~~~$\rightarrow$~~~ & impervious surface \\
river, sea lake & ~~~$\rightarrow$~~~ & water, wetland \\
\bottomrule
\end{tabular}
\end{table}

\subsection{Transfer Model}
\begin{figure}[H]
\begin{subfigure}{.5\textwidth}
  \centering
  \includegraphics[width=.88\linewidth]{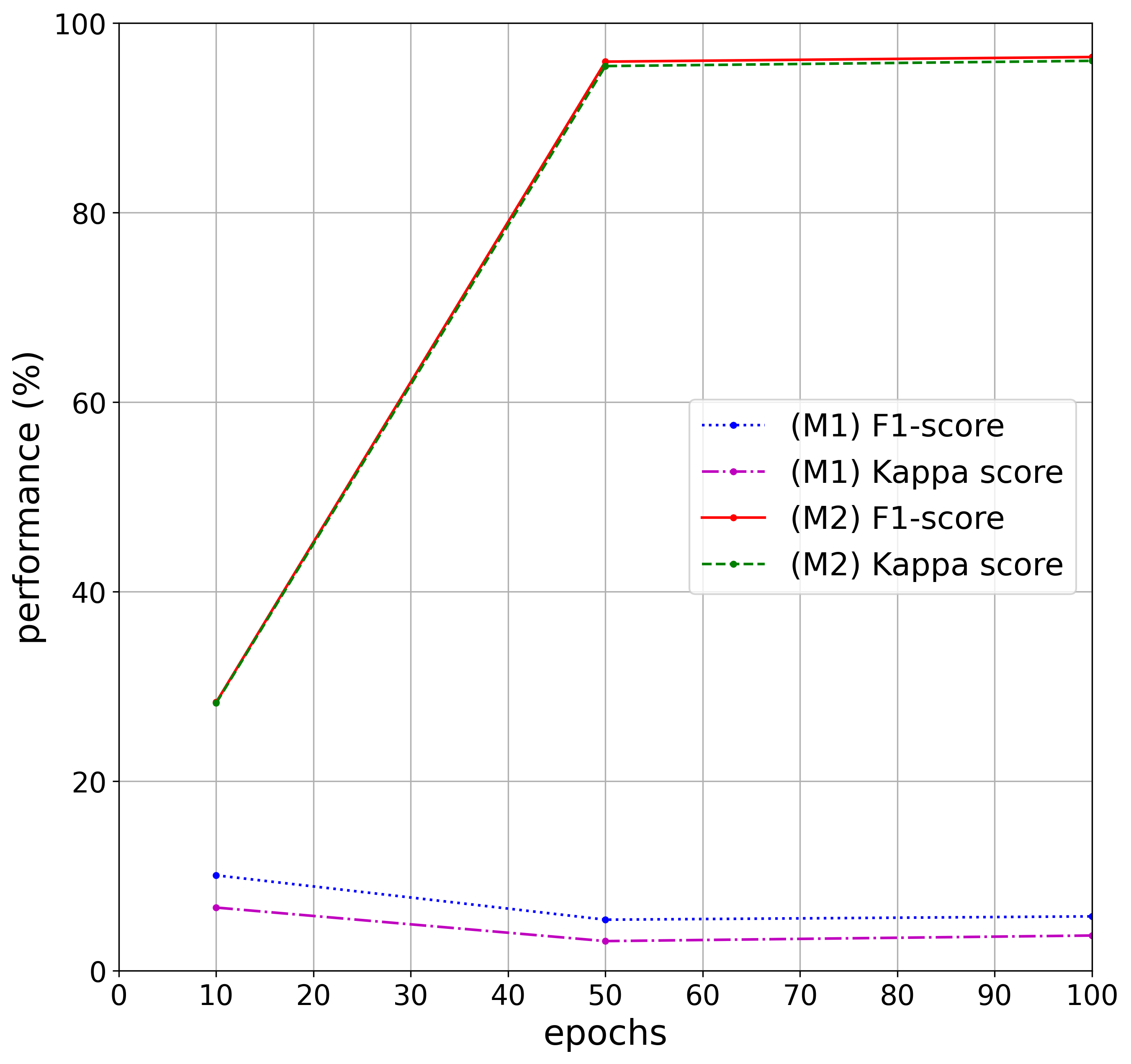}
  \caption{}
  \label{fig:1a}
\end{subfigure}
\begin{subfigure}{.5\textwidth}
  \centering
  \includegraphics[width=.88\linewidth]{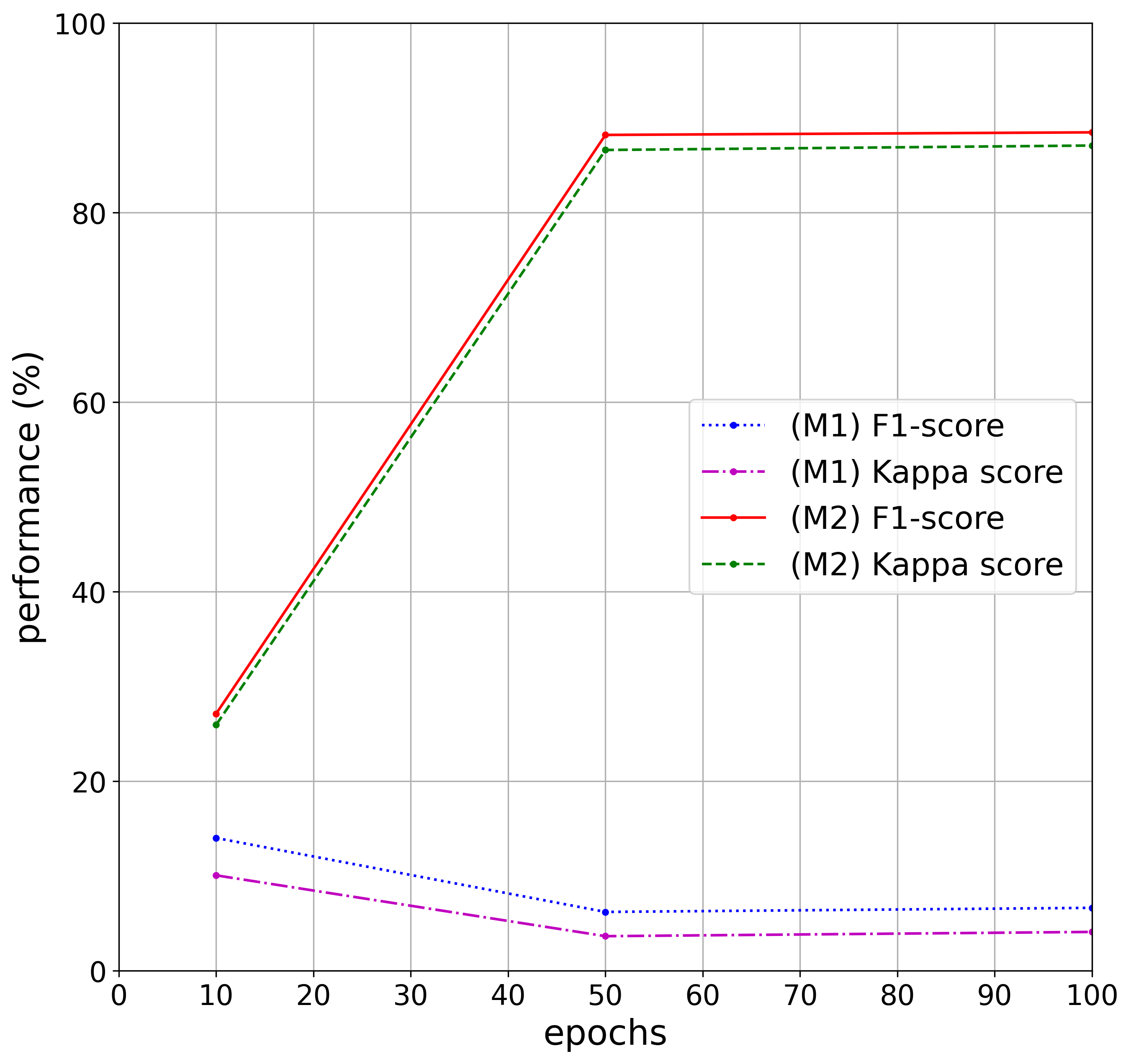}
  \caption{}
  \label{fig:1b}
\end{subfigure}
\caption{Accuracy performance of the adapted \textit{ResNet50} model. M1: 23,587,712 non-trainable and 1,054,233 trainable parameters, and M2: 53,120 non-trainable and 24,591,946 trainable parameters. (\textbf{a})~Both training and testing is based on the {EuroSAT} set; (\textbf{b})~training with the {EuroSAT} set and testing on the {BigEarthNet-v1.0}.}
\label{fig:pretrained}
\end{figure}

To have a more complete picture and with regards to patch-based classification, we show results (Table \ref{tab:resnet50_accuracy} and Figure \ref{fig:pretrained}) of a widely used CNN model, the \textit{ResNet50} \cite{hekaiming2015} with the following two~adaptations:
\begin{itemize}
\item M1: We kept frozen the main topology of the model and we added one 2D convolutional layer at the beginning of the model that merges the four-band input tensor to a standard RGB tensor, and~two dense layers (512 and 10 nodes, respectively) at the end of the model's architecture, resulting in 23,587,712 non-trainable parameters and 1,054,233 trainable parameters. The non-trainable parameters have been tuned based on the \textit{ImageNet} \cite{imagenet2009} data set.
\item M2: We adjusted the \textit{ResNet50} topology in such a way to accept a four-band input tensor and, as in M1, we added two dense layers (512 and 10 nodes, respectively) at the end of the model's architecture, resulting in a new model with 53,120 non-trainable and 24,591,946 trainable parameters. The parameters of the model obtain as initial value the optimal values based on the \textit{ImageNet} training.
\end{itemize}

\begin{table}[H]
\centering
\caption{Classification accuracy of the adapted \textit{ResNet50} models M1 and M2 (100 epochs of training).}
\label{tab:resnet50_accuracy}
\scalebox{0.82}[0.82]{\begin{tabular}{cccllcc}
\toprule
 \textbf{Model} & \textbf{Non-Trainable Parameters} & \textbf{Trainable Parameters} & \textbf{Training} & \textbf{Testing} & \textbf{{F1-Score (\%)}} & \textbf{{Kappa Score (\%)}} \\
\midrule
M1 & 23,587,712 & 1,054,233 & {EuroSAT} & {EuroSAT} & 5.74 & 3.72 \\
M1 & 23,587,712 & 1,054,233 & {EuroSAT} & {BigEarthNet-v1.0} & 6.64 & 4.10 \\
M2 & 53,120 & 24,591,946 & {EuroSAT} & {EuroSAT} & 96.42 & 96.02 \\
M2 & 53,120 & 24,591,946 & {EuroSAT} & {BigEarthNet-v1.0} & 88.47 & 87.08 \\
\bottomrule
\end{tabular}}
\end{table}

\subsection{Data Augmentation}
So far, the data augmentation has been derived by the blending of two different training sets. In this section, we investigate whether there is a gain in terms of accuracy by employing the more conventional data augmentation by means of standard image transformations. Since the image patches of size $5\times5$ are already quite small and their amount reaches very high numbers, we applied the image transformations over the $64\times64$ image patches solely. More specifically, we applied step-wise rotation of step $90^{\circ}$, and flipping in the left-right and up-down direction, resulting in 162,000 samples from the initial 27,000 of {EuroSat} only, and in 192,300 samples from the initial 32,050 of the mixed {EuroSat} \& {BigEarthNet-v1.0} data set. The results are summarised in Table \ref{tab:dataaug}.

\setlength{\tabcolsep}{4pt}
\begin{table}[H]
\centering
\caption{Accuracy performance of CNN-class and adapted {ResNet50} (model M2) on the 10-class patch-based classification problem, computed via tenfold cross-validation, and using data augmentation based on image transformations. The training, validation, and testing sets have been partitioned according to the rule 80/10/10.}
\label{tab:dataaug}
\begin{tabular}{cllcc}
\toprule
 \textbf{Model} & \textbf{Training Set} & \textbf{Testing Set} & {\textbf{F1-Score (\%)}} & {\textbf{Kappa Score (\%)}} \\
\midrule
CNN-class & {EuroSAT} & {EuroSAT} & 99.34 & 99.27 \\
CNN-class & {EuroSAT} & {EuroSAT} \& {BigEarthNet-v1.0} & 89.62 & 88.13 \\
CNN-class & {EuroSAT} \& {BigEarthNet-v1.0} & {EuroSAT} & 99.87 & 99.86 \\
CNN-class & {EuroSAT} \& {BigEarthNet-v1.0} & {EuroSAT} \& {BigEarthNet-v1.0} & 97.08 & 96.73 \\
M2 & {EuroSAT} & {EuroSAT} & 98.07 & 97.85 \\
M2 & {EuroSAT} & {EuroSAT} \& {BigEarthNet-v1.0} & 88.32 & 86.22 \\
M2 & {EuroSAT} \& {BigEarthNet-v1.0} & {EuroSAT} & 99.01 & 98.90 \\
M2 & {EuroSAT} \& {BigEarthNet-v1.0} & {EuroSAT} \& {BigEarthNet-v1.0} & 96.20 & 95.74 \\
\bottomrule
\end{tabular}
\end{table}

\subsection{Experimental Findings}
One of the objectives of the experimental study was to validate whether the presence of bigger amount of representative training samples impact the classification performance comparably or even better than the utilisation of pre-trained and adapted sophisticated models. To verify this hypothesis, we did not use a transfer learning approach in terms of pre-trained and pre-configured NN layers, but~instead, we designed relatively light CNN topologies, keeping the number of model weights as low as possible while retaining the structural capacity of the model in adequate levels. The maximum tenfold cross-validation overall accuracy reaches 99.37\% for CNN-class and 95.41\% for the 3-class output of the CNN-dual. One remark here is that the same image patches shifted for several pixels may produce different results but always in line with the most dominant class. We do not report the accuracy results of CNN-segm because they are slightly worse than the results given by CNN-dual. Nevertheless, the role of CNN-segm is to segment the patches already classified by the CNN-class model in order to refine further the classification (3-class). CNN-dual has more parameters than CNN-segm and has been designed to provide concurrently a 10-class and 3-class segmentation of the image. Both CNN-dual and CNN-segm have been trained on 15,836,588 samples and validated on 3,959,147 samples.

The summarised results of Table \ref{tab:res_accuracy} justify comparatively the gain from the mixing of the training sets in terms of accuracy and robustness. The two measures, F1 and Kappa score, improve significantly when the training is based on both data sets than employing a single one (the {EuroSAT} in this case).

The results shown in Figure \ref{fig:confmatrix} corroborate the confusion between the ``green'' vegetation classes that is due to the seasonality effect. The fact that several \textit{cropland} and \textit{grassland} pixels have been classified to the \textit{impervious} class can be explained by the fact that pixels reflecting rural roads or small settlements pull their surrounding pixels as well in many cases, resulting in thicker objects. This~phenomenon is strongly dependent on the detection capacity of the sensor. Nevertheless, the~agreement between the two layers is noteworthy if we consider that the selection of the $5\times5$ image patches have not undergone a refinement in terms of class labelling. More precisely, as the original $64\times64$ image patches have been divided into $5\times5$ blocks, likewise, the class label of each $64\times64$ image patch has been assigned to all its constituent  $5\times5$ blocks, causing several false class assignments since there are more than one classes in most of the $64\times64$ image patches. Currently, we~are working in this front, trying to find an automatic way for the precise assignment of class labels to the smaller blocks, something that will improve a lot the classification results.

The very low classification performance shown in Table \ref{tab:resnet50_accuracy} with respect to model (M1) states that this type of transfer learning is not adequate at all since the low number of trainable parameters is not sufficient to help the model adapt into the new domain. In regard to the model (M2), the~same Figures show that the classification performance is similar to the results of the patch-based classification (CNN-class) as presented in Table \ref{tab:res_accuracy}, especially in the case of 100 epochs. Nevertheless, the lower number of parameters of CNN-class, adjusted from scratch upon the combined {EuroSAT} and {BigEarthNet-v1.0} training set seems more adequate option in terms of computational complexity and adaptability than using a more complex model trained in one of the two sets.

The positive effect of data augmentation through standard image transformations is apparent as shown by the results of Table \ref{tab:dataaug}. However, if we compare the pair \textit{training: {BigEarthNet-v1.0} \& {EuroSAT}} and \textit{testing: {EuroSAT}} of Table \ref{tab:res_accuracy} with the pair \textit{training: {EuroSAT}} and \textit{testing: {BigEarthNet-v1.0} \& {EuroSAT}} of Table \ref{tab:dataaug}, we acknowledge the fact that data augmentation via image transformations alone, when applied on {EuroSAT}, does not improve the classification result so much compared to the data augmentation that has been derived by the blending of the two training sets {BigEarthNet-v1.0} \& {EuroSAT}.

We do not provide comparable results of transfer learning for image segmentation because, as we have already mentioned, the original purpose of both {EuroSAT} and {BigEarthNet-v1.0} data sets was the patch-based classification, and the only way to convert the problem into a semantic segmentation problem is to deal with image blocks of size much smaller than the $64\times64$ patch size supported by the data sets. Although some of the standard pre-trained segmentation models can operate with the original $64\times64$ image blocks for the 3-class image segmentation, they cannot answer on the 10-class image segmentation problem due to the lack of detailed reference masks. This challenging problem has been tackled by our customised CNN model (CNN-dual) which operates at pixel level and can be trained fast from scratch upon the fused data set. It supports as well our hypothesis that the existence of big enough, representative, and enriched data sets combined with customised modelling approaches provides added-value to the classification outcome compared to transfer learning.


\section{Open Access to Data and Workflows}
\label{sec:comp_platform}
Training deep neural networks requires hardware and software libraries to be fine-tuned for array-based intensive computations. Multi-layered networks rely heavily on matrix math operations and demand immense amounts of computing capacity (mostly floating-point). For some years now, the state of the art in such type of computing and especially for image processing is shaped by powerful machinery such as the graphics processing units (GPUs) and their optimised architectures.

In this regard, the JRC (Joint Research Centre) Big Data Analytics project, having as a major objective to provide services for large-scale processing and data analysis to the JRC scientific community and the collaborative partners, is constantly increasing the fleet of GPU-based processing nodes, including NVIDIA$^{\textregistered}$ Tesla K80, GeForce GTX 1080 Ti, Quadro RTX 6000, and Tesla V100-PCIE cards. Dedicated Docker images with CUDA \cite{Whitehead2018} parallel model, back-end, and deep learning frameworks such as TensorFlow, Apache MXNet and PyTorch \cite{Paszke2019} and adequate application programming interfaces have been configured to facilitate and streamline the prototyping and large-scale testing of working models.

We mention these technical details in order to underline the fact that although operations such as transfer learning, domain adaptation, model customization, and hyper-parameter fine-tuning are much lighter than training and optimising a deep neural network from scratch, they also require dedicated hardware and software for the exploration of various scenarios and the shortening of the experimentation process. 

The entire experimental setting presented here has been performed onto the JRC's high-throughput computing platform, the so-called JEODPP \cite{Soille2018}. Jupyter notebooks and Docker images are open and accessible upon request. This decision is in conformity with the {FAIR} Guiding Principles for scientific data management and stewardship, and promotes open science. Complete or sectional download of the {SatImNet} collection can be done via \textit{ftp
} (\mbox{\url{https://jeodpp.jrc.ec.europa.eu/ftp/public/MachineLearning/SatImNet}}) service. In addition, individual files are directly accessible through the support of \textit{vsizip} and \textit{vsicurl} drivers. The open repository (\mbox{\url{https://github.com/syrriva/SatImNet}}) contains the Python scripts (in the form of Jupyter notebooks) to access and query the SatImNet collection via \textit{ftp}.

\section{Conclusions}
\label{sec:conclusion}
The availability and plurality of well-organised, complete, and representative data sets is critical for the efficient training of machine learning models (specifically of deep neural networks) in order to solve satellite image classification tasks in a robust and operative fashion. Working under the framework of open science and very closely to policy support which invites for transparent and reproducible processing pipelines at which data and software are integrated, are open and freely available through ready-to-use working environments, the contribution of this paper is aligned with three goals: (i) to define the functional characteristics of a sustainable collection of training sets, aiming at covering the various specificities that delineate the landscape of automated satellite image classification; (ii) according to the defined attributes, to structure and compile a number of heterogeneous data sets, and (iii) to demonstrate a potential fusion of training sets by using deep neural network modelling and solving concurrently an image classification and segmentation problem.

Future work involves systematic harvesting of training sets across Internet, automation of the quality control of the discovered data sets, and continuous integration of the distinct modules of the working pipeline. Apart from the accumulation of data sets which have been designed and provided by the research community, another scheduled activity concerns the methodical building of in-house training sets, targeting wide scope Earth observation applications such as crop and surface water monitoring, deforestation, and crisis management.
\vspace{6pt}

\newpage
\authorcontributions{V.S. and P.S. planned the formulation of the SatImNet data set; V.S. and O.P. selected the open data and O.P. downloaded the data sets in their original version; V.S., O.P., and P.S. defined the attributes on which the data sets have been categorised; V.S. designed and implemented the experimental study, and structured the data; V.S. created all the Jupyter notebooks and O.P. tested them; V.S. wrote the paper and all the authors reviewed and edited it. All authors have read and agreed to the published version of the manuscript.}
\funding{This research received no external funding.} 

\acknowledgments{
The authors would like to thank Tom\'{a}\v{s} Kliment, Panagiotis Mavrogiorgos, Pier Valerio Tognoli, and Paul Hasenohr for their contribution in data management, ftp service set up, and Docker images configuration and maintenance.
}
\conflictsofinterest{The authors declare no conflict of interest.}
\reftitle{References}

\end{document}